\documentclass[10pt,twocolumn,letterpaper]{article}

\usepackage{cvpr}              
\usepackage{multirow}
\usepackage{algorithm}
\usepackage{algpseudocode}
\usepackage{multicol}
\usepackage{listings}

\definecolor{cvprblue}{rgb}{0.21,0.49,0.74}
\usepackage[pagebackref,breaklinks,colorlinks,allcolors=cvprblue]{hyperref}

\def\paperID{13744} 
\def\confName{CVPR}
\def\confYear{2025}

\definecolor{my_pink}{HTML}{be0027}

\let\oldtwocolumn\twocolumn
\renewcommand\twocolumn[1][]{
    \oldtwocolumn[{#1}{
        \centering
        \vspace{-20pt}
        \includegraphics[width=\textwidth]{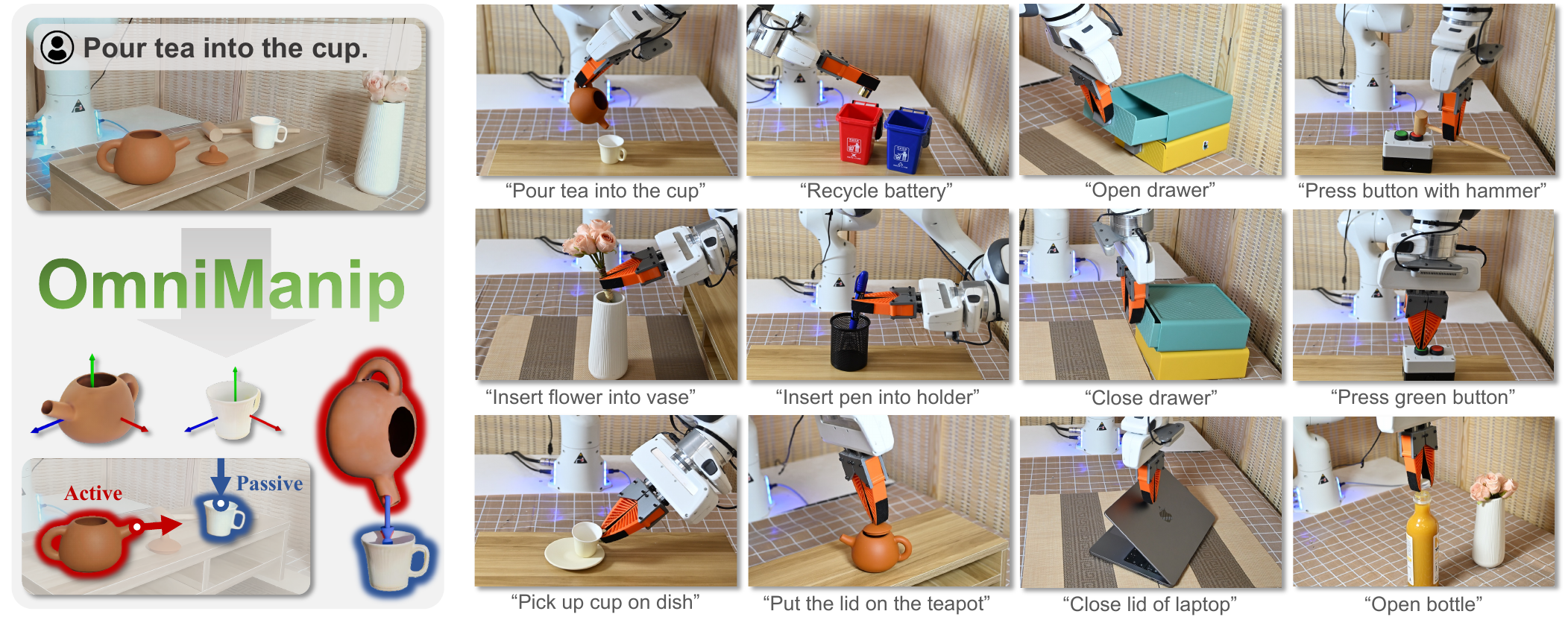}  
        \vspace{-15pt}
        \captionof{figure}{We proposed \textbf{OmniManip}, an open-vocabulary manipulation method that bridges the gap between the high-level reasoning of vision-language models (VLM) and the low-level precision, featuring closed-loop capabilities in both planning and execution.
        }      
        \vspace{5pt}
        \label{fig:teaser}
    }]
}

\title{OmniManip: Towards General Robotic Manipulation via Object-Centric Interaction Primitives as Spatial Constraints}

\author{
    Mingjie Pan$^{1,2*}$, 
    Jiyao Zhang$^{1,2*}$, 
    Tianshu Wu\textsuperscript{\rm 1},
    Yinghao Zhao\textsuperscript{\rm 3}, 
    Wenlong Gao\textsuperscript{\rm 3},
    Hao Dong$^{1,2\dagger}$ \vspace{2pt}\\
    \textsuperscript{\rm 1}CFCS, School of CS, Peking University \quad 
    \textsuperscript{\rm 2}PKU-AgiBot Lab  \quad 
    \textsuperscript{\rm 3}AgiBot \vspace{2pt} \\
    \href{https://omnimanip.github.io/}{\textcolor{my_pink}{https://omnimanip.github.io}}
}

\def\eg{\emph{e.g}.}

\newcommand{\Tref}[1]{Table~\ref{#1}}
\newcommand{\Fref}[1]{Figure~\ref{#1}}
\newcommand{\Eref}[1]{Equation~\ref{#1}}
\newcommand{\Sref}[1]{Section~\ref{#1}}

\begin{document}

\maketitle
\begin{abstract}
\renewcommand{\thefootnote}{}
\footnote{*: Equal contributions. $\dagger$: Corresponding author}
The development of general robotic systems capable of manipulating in unstructured environments is a significant challenge. While Vision-Language Models(VLM)  excel in high-level commonsense reasoning, they lack the fine-grained 3D spatial understanding required for precise manipulation tasks. Fine-tuning VLM on robotic datasets to create Vision-Language-Action Models(VLA) is a potential solution, but it is hindered by high data collection costs and generalization issues. To address these challenges, we propose a novel object-centric representation that bridges the gap between VLM's high-level reasoning and the low-level precision required for manipulation. Our key insight is that an object's canonical space, defined by its functional affordances, provides a structured and semantically meaningful way to describe interaction primitives, such as points and directions. These primitives act as a bridge, translating VLM's commonsense reasoning into actionable 3D spatial constraints. In this context, we introduce a dual closed-loop, open-vocabulary robotic manipulation system: one loop for high-level planning through primitive resampling, interaction rendering and VLM checking, and another for low-level execution via 6D pose tracking. This design ensures robust, real-time control without requiring VLM fine-tuning. Extensive experiments demonstrate strong zero-shot generalization across diverse robotic manipulation tasks, highlighting the potential of this approach for automating large-scale simulation data generation.
\end{abstract}
\vspace{-15pt}
\section{Introduction}
\label{sec:intro}
\vspace{-5pt}
Developing a general robotic manipulation system has long been a challenging task, primarily due to the complexity and variability of real-world~\cite{rgbgrasp, wu2024canonical,graspgf}. Inspired by the rapid advancements in Large Language Models (LLM)\cite{gpt4, llama2} and Vision-Language Models (VLM) \cite{clip, sig_lip, llava, sphinx}, which leverage vast amounts of internet data to acquire rich commonsense knowledge, researchers have recently turned attention to exploring their application in robotics\cite{lvdiffusor,voxposer}. Most existing works focus on utilizing this knowledge for high-level task planning, such as semantic reasoning \cite{progprompt, reflect, spatial_vlm}. 
Despite these advances, current VLMs, primarily trained on extensive 2D visual data, lack the 3D spatial understanding ability necessary for precise, low-level manipulation tasks. This limitation poses challenges in manipulations within unstructured environments.

One approach to overcoming this limitation is to fine-tune VLM on large-scale robotic datasets, transforming them into VLA \cite{rt1,rt2,palm_e,openvla}. However, this faces two major challenges: 1) acquiring diverse, high-quality robotic data is costly and time-consuming, and 2) fine-tuning VLM into VLA results in agent-specific representations, which are tailored to specific robots, limiting their generalizability.
A promising alternative is to abstract robotic actions into interaction primitives (\eg, points or vectors) and leverage VLM reasoning to define the spatial constraints of these primitives, while traditional planning algorithms handle execution \cite{copa, rekep, moka}. However, existing methods for defining and using primitives have several limitations: The process of generating primitive proposals is task-agnostic, which poses the risk of lacking suitable proposals. Additionally, relying on manually designed rules for post-processing proposals also introduces instability.
This naturally leads to an important question: 
\textit{How can we develop more efficient and generalizable representations that bridge VLM high-level reasoning with precise, low-level robotic manipulation?}

To address this challenge, we propose a novel object-centric intermediate representation incorporating interaction points and directions within an object's canonical space. This representation bridges the gap between VLM's high-level commonsense reasoning and precise 3D spatial understanding. 
Our key insight is that an object's canonical space is typically defined based on its functional affordances. As a result, we can describe an object's functionality in a more structured and semantically meaningful way within its canonical space. 
Meanwhile, recent advancements in universal object pose estimation~\cite{dreds,genpose,omni6dpose} make it feasible to canonicalize a wide range of objects.

Specifically, we employ a universal 6D object pose estimation model \cite{omni6dpose} to canonicalize objects and describe their rigid transformations during interactions. In parallel, a single-view 3D generation network generates detailed object meshes \cite{triposr, One2345++}.
Within the canonical space, interaction directions are initially sampled along the object’s principal axes, providing a coarse set of interaction possibilities. Meanwhile, the VLM predicts interaction points. Subsequently, the VLM identifies task-relevant primitives and estimates the spatial constraints between them. 
To address the hallucination issue in VLM reasoning, we introduce a self-correction mechanism through interaction rendering and primitive resampling that enables closed-loop reasoning. Once the final strategy is determined, actions are computed through constrained optimization, with pose tracking ensuring robust, real-time control in a closed-loop execution phase.
Our method offers several key advantages: 
1) \textbf{Efficient and Effective Interaction Primitive Sampling}: By leveraging the object's canonical space, our approach enables efficient and effective sampling of interaction primitives, enhancing the system’s reasoning capabilities.
2) \textbf{Dual Closed-Loop, Open-Vocabulary Robotic Manipulation System}: Benefiting from the proposed object-centric intermediate representation, our method implements a dual closed-loop system. The rendering and resampling process drives a reasoning loop for decision-making, while pose tracking ensures a closed loop for action execution. 

In summary, our contributions are threefold: 
\begin{itemize}
\item We propose a novel object-centric interaction representation that bridges the gap between VLM's high-level commonsense reasoning and low-level robotic manipulation.
\item To the best of our knowledge, we are the first to present a planning and execution dual closed-loop open-vocabulary manipulation system without VLM fine-tuning.
\item Extensive experiments demonstrate our method's strong zero-shot generalization across diverse manipulation tasks, and we also highlight its potential for automating robotic manipulation data generation.
\end{itemize}

\vspace{-3pt}
\section{Related Work}
\label{sec:related_work}
\vspace{-3pt}

\begin{figure*}[t]
\centering
\includegraphics[width=\linewidth]{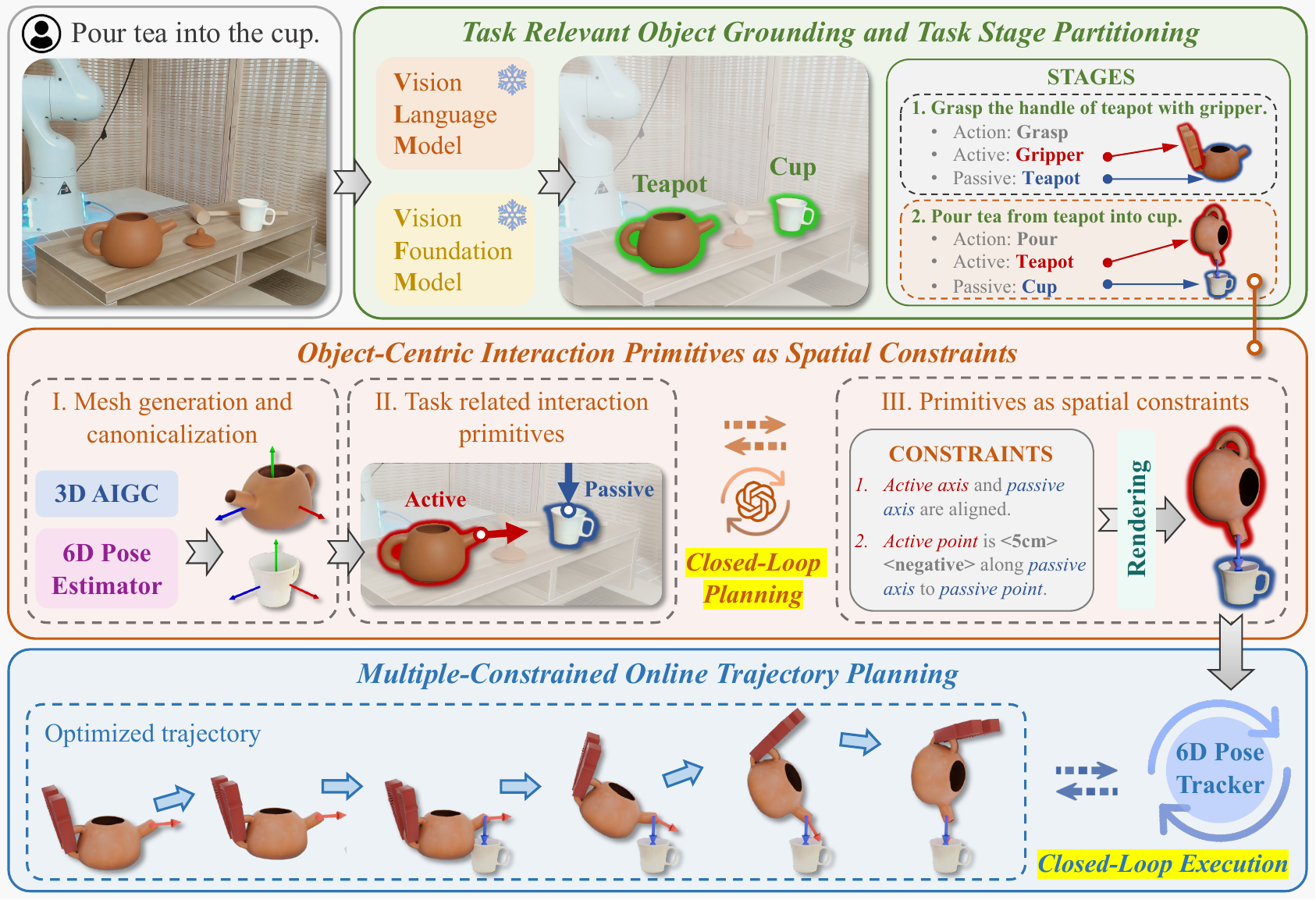}
\vspace{-20pt}
\caption{
\textbf{Overview framework.}
Given instruction and RGB-D observation marked by VFM, VLM firstly filters task-related objects and partitions the task into stages. For each stage, VLM extracts object-centric canonical interaction primitives as spatial constraints in a closed-loop manner. For execution, the trajectory is optimized by constraints and updated in a closed loop using a 6D Pose Tracker.
}
\label{fig:method_overview} 
\vspace{-16pt}
\end{figure*}

\noindent\textbf{Foundation Models For Robotics}
The emergence of foundation models has significantly influenced the field of robotics\cite{foundation_for_robot_survey1, foundation_for_robot_survey2, foundation_for_robot_survey3}, particularly in the application of vision-language models\cite{gpt4, llava, blip2, lidarllm, 3dllm,  spatial_vlm}, which excel in environment understanding and high-level commonsense reasoning. 
These models demonstrate the potential for controlling robots to perform general tasks in novel and unstructured environments. 
Some studies \cite{rt1, rt2, roboflamingo, openvla} have fine-tuned VLM on robotics datasets to create VLA models that output robotic trajectories, but these efforts are limited by the high cost of data collection and issues with generalization. 
Other approaches attempt to extract operation primitives using visual foundation models \cite{pivot, moka, copa, rekep, manipulate_anything, robopoint, affordance_visual_prompt}, which are then used as visual or language prompts for VLM to perform high-level commonsense reasoning, combined with motion planners \cite{ompl, sequence_mpc, curobo} for low-level control. 
However, these methods are constrained by the ambiguity of compressing 3D primitives into the 2D images or 1D text required by VLM and the hallucination tendencies of VLM themselves, making it difficult to ensure that the high-level plans generated by VLM are accurate. 
In this work, we demonstrate OmniManip's unique advantages in addressing these challenges, particularly in fine-grained 3D understanding and mitigating large model hallucinations.

\noindent\textbf{Representations for Manipulation}
Structural representations determine the capabilities and effectiveness of manipulation methods. Among various types of representations,  keypoints are a popular choice due to their flexibility, generalization, and ability to model variability \cite{bad_points_1,bad_points_2,bad_points_3,atm}. However, these keypoints-based methods require manual task-specific annotations to generate actions. To enable zero-shot open-world manipulation, studies such as \cite{rekep, moka, pivot} have transformed keypoints into visual prompts for VLM, facilitating the automatic generation of high-level planning results. Despite their advantages, keypoints can be unstable; they struggle under occlusion and pose challenges in the extraction and selection of specific keypoints.
Another common representation is the 6D pose, which efficiently defines long-range dependencies between objects for manipulation and offers a degree of robustness to occlusion \cite{bad_pose_1, bad_pose_2, bad_pose_3,foundationpose}. However, these methods necessitate prior modeling of geometric relationships and, due to the sparse nature of poses, cannot provide fine-grained geometry. This limitation can lead to failures in manipulation strategies across different objects due to intra-class variations.
To address these issues, OmniManip combines the fine-grained geometry of keypoints with the stability of the 6D pose. It automatically extracts detailed functional points and directions within the canonical coordinate system of objects using VLM, enabling precise manipulation.

\section{Method}
\label{sec:method}

Here we discuss: 
\textbf{(1)} How do we formulate robotic manipulation via interaction primitives as spatial constraints(Sec. \ref{sec:formulation})? 
\textbf{(2)} How to extract canonical interaction primitives in a generic and open vocabulary way (Sec. \ref{sec:primitives})?
\textbf{(3)} Why can OmniManip achieve a dual closed-loop system (Sec. \ref{sec:dual_close_loop})?

\subsection{Manipulation with Interaction Primitives}
\label{sec:formulation}
In our formulation, complex robotic tasks are decomposed into stages, each defined by object interaction primitives with spatial constraints. This structured approach allows for the precise definition of task requirements and facilitates the execution of complex manipulation tasks. In this section, we detail how interaction primitives serve as the foundation for spatial constraints, enabling robust manipulation.

\noindent\textbf{Task Decomposition.}  
As shown in~\Fref{fig:method_overview}, given a manipulation task \( \mathcal{T} \) (\eg, \emph{pouring tea into a cup}), we first utilize GroundingDINO\cite{grounding_dino} and SAM\cite{sam}, two Visual Foundation Models (VFMs),  to mark all foreground objects in the scene like \cite{som} as visual prompt.  Subsequently, a VLM \cite{gpt4} is employed to filter task-relevant objects and decompose the task into multiple stages \( \mathcal{S} = \{\mathcal{S}_1, \mathcal{S}_2, \dots, \mathcal{S}_n\} \), where each stage \( \mathcal{S}_i \) can be formalized as \( \mathcal{S}_i = \{A_i, \mathcal{O}_i^{\text{active}}, \mathcal{O}_i^{\text{passive}}\} \), where \( A_i \) represents the action to be performed (\eg, grasp, pour), and \( \mathcal{O}_i^{\text{active}} \) and \( \mathcal{O}_i^{\text{passive}} \) refer to the object initiating the interaction and the object being acted upon, respectively.
For example, in~\Fref{fig:method_overview}, the teapot is the passive object in the stage of grasping the teapot while the teapot is the active object and the cup is passive in the stage of pouring tea into the cup.

\noindent\textbf{Object-Centric Canonical Interaction Primitives.}
We propose a novel object-centric representation with canonical interaction primitives to describe how objects interact during manipulation tasks. 
Specifically, an object's interaction primitives are characterized by its interaction point and direction in canonical space. The interaction point \( \mathbf{p} \in \mathbb{R}^3 \) denotes a key location on the object where interaction occurs, while the interaction direction \( \mathbf{v} \in \mathbb{R}^3 \) represents the primary axis relevant to the task. Together, these form the interaction primitive \( \mathcal{O} = \{\mathbf{p}, \mathbf{v}\} \), encapsulating the essential intrinsic geometric and functional properties required to meet task constraints.
These canonical interaction primitives are defined relative to their canonical space, remaining consistent across different scenarios, enabling more generalized and reusable manipulation strategies.

\begin{figure}[t]
    \centering
    \includegraphics[width=\linewidth]{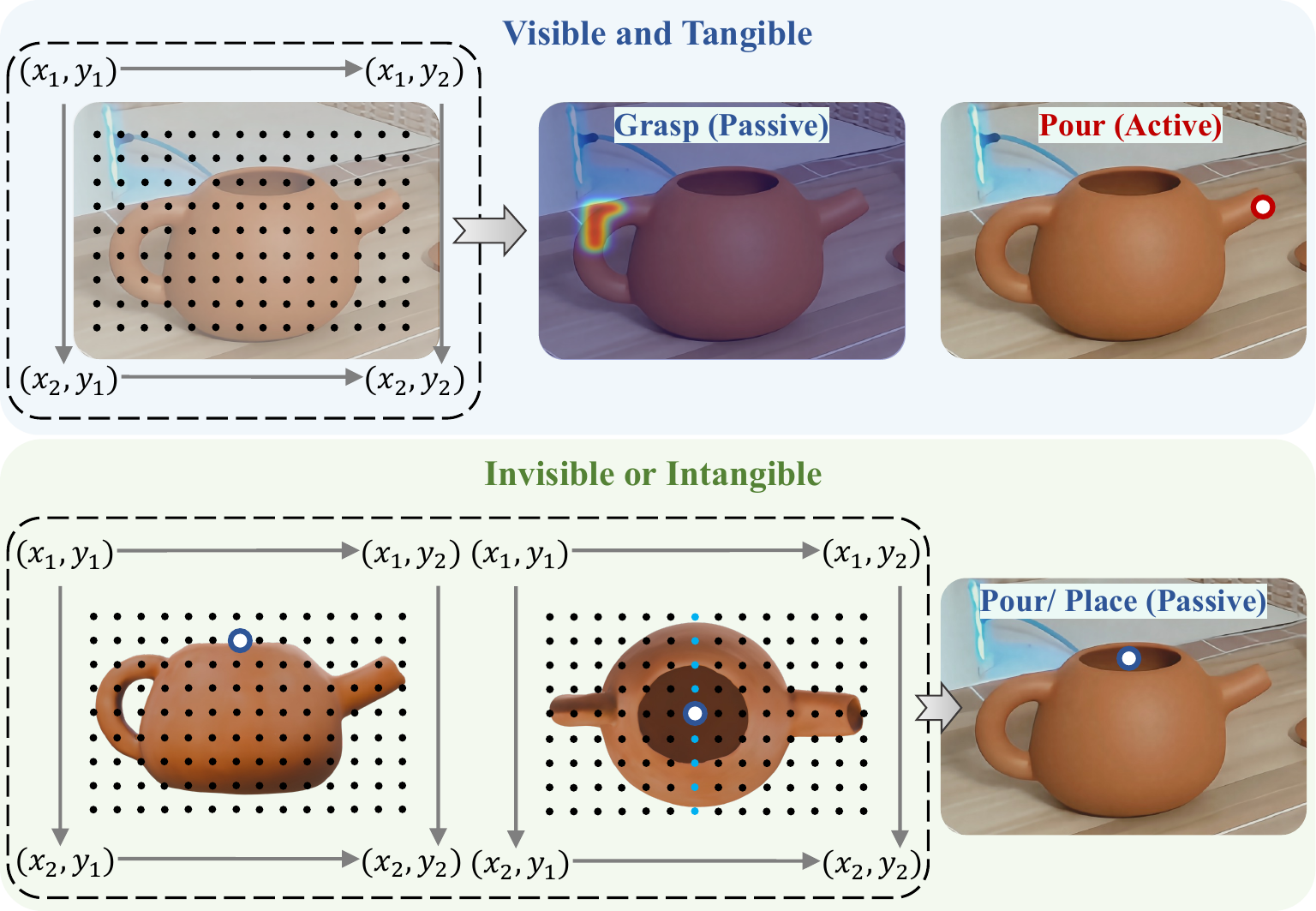} 
    \vspace{-20pt}
    \caption{Interaction points generation.} 
    \label{fig:primitives_points}
    \vspace{-20pt}
\end{figure}

\noindent\textbf{Interaction Primitives with Spatial Constraints.} At each stage \( \mathcal{S}_i \), a set of spatial constraints \( \mathcal{C}_i \) governs the spatial relationships between the active and passive objects. These constraints are divided into two categories: \textit{distance constraints} \( d_i \), which regulate the distance between interaction points, and \textit{angular constraints} \( \theta_i \), which ensure proper alignment of interaction directions. Together, these constraints define the geometric rules necessary for precise spatial alignment and task execution. The overall spatial constraint for each stage \( \mathcal{S}_i \) is given by:

\begin{equation}
\mathcal{C}_i = \left\{ \mathcal{O}_i^{\text{active}}, \mathcal{O}_i^{\text{passive}}, d_i, \theta_i \right\}
\label{equation:constraints}
\end{equation}

Once the constraints \( \mathcal{C}_i \) have been defined, the task execution can be formulated as an optimization problem. 


\begin{figure}[t]
    \centering
    \includegraphics[width=\linewidth]{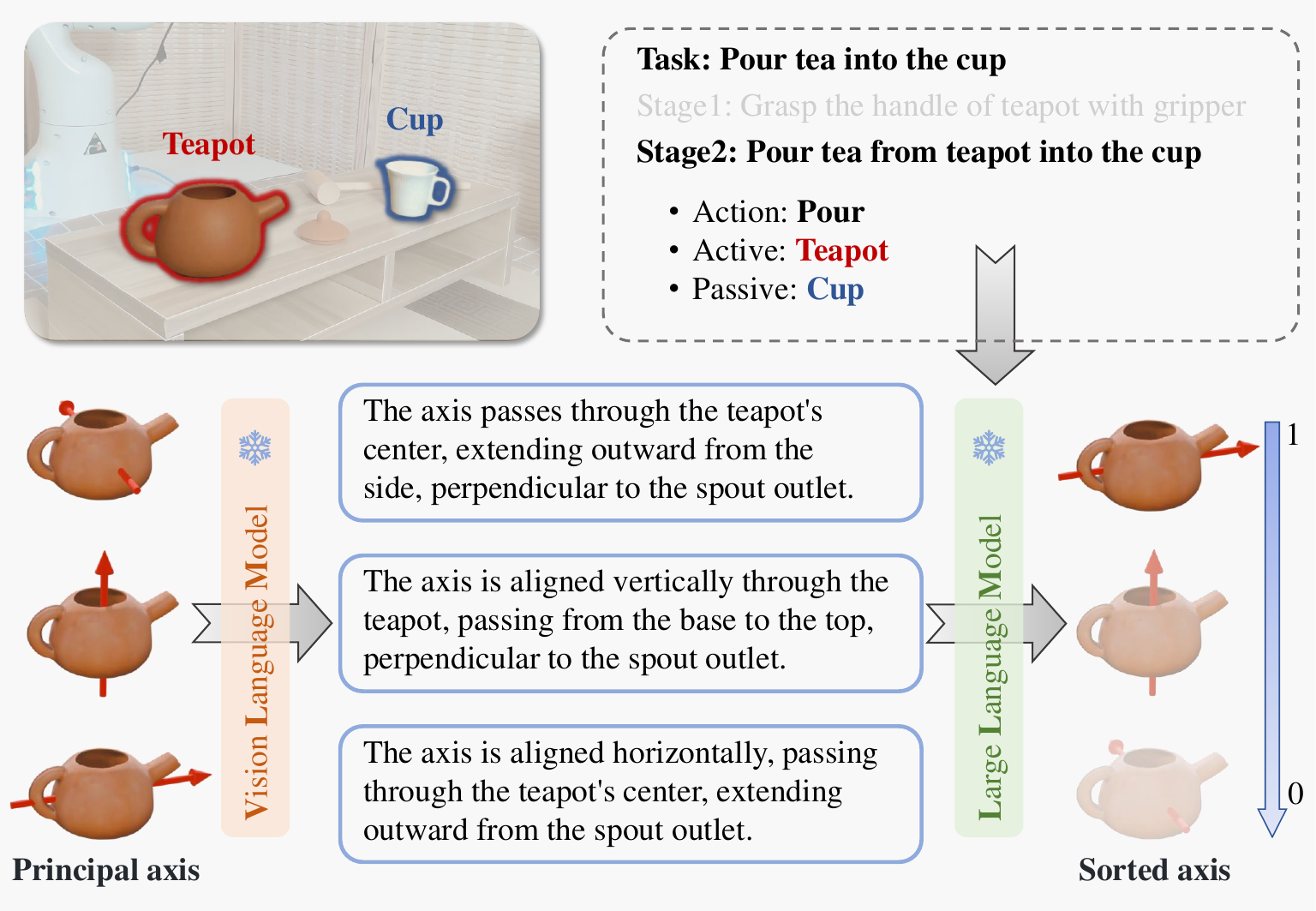} 
    \vspace{-20pt}
    \caption{Interaction directions extraction.}
    \label{fig:primitives_directions}
    \vspace{-20pt}
\end{figure}

\subsection{Primitives and Constraints Extraction}
\label{sec:primitives}

In this section, we detail the process of extracting interaction primitives and their spatial constraints \(\mathcal{C}\) for each stage. As illustrated in~\Fref{fig:method_overview}, we first obtain 3D object meshes for both the task-relevant active and passive objects via single-view 3D generation~\cite{triposr, zou2024triplane, One2345++}, followed by pose estimation with Omni6DPose\cite{omni6dpose} for object canonicalization. Next, we extract task-relevant interaction primitives and their corresponding constraints.

\noindent\textbf{Grounding Interaction Point.}  
As shown in~\Fref{fig:primitives_points}, interaction points are categorized as \textit{Visible and Tangible} (\eg, a teapot handle) or \textit{Invisible or Intangible} (\eg, the center of its opening). To enhance VLM for interaction points grounding, SCAFFOLD~\cite{scaffold} visual prompting mechanism is employed, which overlays a Cartesian grid onto the input image. Visible points are directly localized in the image plane, while invisible points are inferred through multi-view reasoning based on proposed canonical object representations, as illustrated in~\Fref{fig:primitives_points}. Reasoning begins from the primary viewpoint, with ambiguities resolved by switching to an orthogonal view. This approach enables more flexible and reliable interaction point grounding. For tasks like grasping, heatmaps are generated from multiple interaction points, improving the robustness of the grasping model.

\noindent\textbf{Sampling Interaction Direction.}
In the canonical space, the principal axes of an object are often functionally relevant. As illustrated in~\Fref{fig:primitives_directions}, we treat the principal axes as candidate interaction directions. However, assessing the relevance of these directions to the task is challenging due to the limited spatial understanding of the current VLM. To address this, we propose a VLM caption and LLM scoring mechanism: first, we use the VLM to generate semantic descriptions for each candidate axis, and then employ a LLM to infer and score the relevance of these descriptions to the task. This process results in an ordered set of candidate directions that are most aligned with the task requirements.

Ultimately, the interaction primitives with constraints are generated with VLM, yielding an ordered list of constrained interaction primitives for each stage \( \mathcal{S}_i \), denoted as \(K_i = \{C_i^{(1)}, C_i^{(2)}, \dots, C_i^{(N)}\}\).

\begin{table*}[t]
\centering
\setlength\tabcolsep{15pt}
\begin{tabular}{lccccc}
\toprule
\textbf{Tasks} & \textbf{VoxPoser} &\textbf{CoPa} &  \textbf{ReKep}& \multicolumn{2}{c}{\textbf{OmniManip(Ours)}} \\ 
\cline{5-6}
 &  &  & Auto & Closed-loop & Open-loop \\ \midrule \midrule
Pour tea                    & 0/10 & 1/10 & 3/10 & \textbf{7/10} & 6/10  \\
Insert flower into vase     & 0/10 & 4/10 & 2/10 & \textbf{6/10} & 4/10  \\   
Insert the pen in holder    & 0/10 & 4/10 & 3/10 & \textbf{7/10} & 5/10  \\   
Recycle the battery         & 6/10 & 5/10 & 7/10 & \textbf{8/10} & 6/10  \\   
Pick up the cup on the dish & 3/10 & 2/10 & \textbf{9/10} & 8/10 & 7/10  \\   
Fit the lid onto the teapot & 0/10 & 2/10 & 3/10 & \textbf{5/10} &  3/10 \\
\midrule
\textbf{Total} & 15.0\% & 30.0\% & 45.0\% & \textbf{68.3\%} &  51.7\% \\ \midrule

Open the drawer             & 1/10 & 4/10 &  -   & \textbf{6/10} &  4/10 \\  
Close the drawer            & 3/10 & 3/10 &  -   & \textbf{8/10} & 6/10  \\  
Hammer the button           & 0/10 & 3/10 &  -   & \textbf{4/10} & 2/10  \\  
Press the red button        & 0/10 & 3/10 &  -   & \textbf{7/10} &  6/10 \\   
Close the lid of the laptop & 4/10 & 3/10 &  -   & \textbf{6/10} &  4/10 \\  
Open the jar                & 2/10 & 0/10 &  -   & \textbf{6/10} & 5/10  \\  
\midrule
\textbf{Total}              &   16.7\%   &  26.7\%    &   -   &     \textbf{61.7\%}    &  45.0\%  \\  \bottomrule

\end{tabular}

\vspace{-4pt}
\caption{\textbf{Quantitative results across 12 real-world manipulation tasks.} The first six tasks focus on rigid object manipulation, while the latter involves articulated object manipulation. `-' indicates that the method can not handle this task due to its underlying principles.
 }
\label{table:main_table}
\vspace{-16pt}
\end{table*}

\subsection{Dual Closed-Loop System}
\label{sec:dual_close_loop}
As outlined in~\Sref{sec:primitives}, we obtain the interaction primitives of the active and passive objects, denoted as \( \mathcal{O}^{\text{active}} \) and \( \mathcal{O}^{\text{passive}} \), respectively, along with the spatial constraints \( \mathcal{C} \) that define their spatial relationships. However, this is an open-loop inference, which inherently limits the robustness and adaptability of the system. These limitations arise primarily from two sources: 1) the hallucination effect in large models, and 2) the dynamic nature of real-world environments.
To overcome these challenges, we propose a dual closed-loop system, as illustrated in~\Fref{fig:method_overview}. 

\begin{algorithm}
\caption{Self-Correction Algorithm via \textbf{RRC}}
\textbf{Input:} Task $\mathcal{T}$, Stage $\mathcal{S}_i$, Initial List of Primitives with Constraints $\mathcal{K}_i = \left\{ \mathcal{C}_i^{(1)}, \mathcal{C}_i^{(2)}, \dots, \mathcal{C}_i^{(N)} \right\}$\\
\textbf{Output:} Successful Constraints $\hat{\mathcal{C}}_i$ or Task Failure
\begin{algorithmic}[1]
\State $k \gets 1$, $maxSteps \gets N$, $refine \gets \textbf{False}$
\While{$k \leq maxSteps$}
    \State $k \gets k + 1$

    \State \textbf{Render:} $\mathbf{I}_i \gets \text{Render}(\mathcal{C}_i^{(k)})$    
    \State \textbf{Check:} $state \gets \text{VLM}(\mathcal{T}, \mathcal{S}_i, \mathbf{I}_i, \mathcal{C}_i^{(k)}, refine)$

    \If{$state = \text{`Refine'}$ \textbf{and} $refine = \textbf{False}$}
        \State \textbf{Resample:} Update $\mathcal{K}_i \gets \text{Resample}(\mathcal{C}_i^{(k)})$
        \State $k \gets 1$, $maxSteps \gets M$, $refine \gets \textbf{True}$
    \ElsIf{$state = \text{`Success'}$}
        \State \Return $\mathcal{C}_i^{(k)}$ 
    \EndIf
\EndWhile
\State \Return \textbf{Task Failed} 
\end{algorithmic}
\end{algorithm}

\noindent\textbf{Closed-loop Planning.}
To improve the accuracy of interaction primitives and mitigate hallucination issues in the VLM, we introduce a self-correction mechanism based on \textbf{R}esampling, \textbf{R}endering, and \textbf{C}hecking (\textbf{RRC}). This mechanism uses real-time feedback from a visual language model (VLM) to detect and correct interaction errors, ensuring precise task execution. The RRC process consists of two stages: the \textit{initial phase} and the \textit{refinement phase}. The overall RRC mechanism is outlined in Algorithm 1.
In the initial phase, the system evaluates the interaction constraints \( \mathcal{K}_i \) defined in~\Sref{sec:primitives}, which specify the spatial relationships between active and passive objects. For each constraint \( \mathcal{C}_i^{(k)} \), the system renders an interaction image \( \mathbf{I}_i \) based on the current configuration and submits it to the VLM for validation. The VLM returns one of three outcomes: \textit{success}, \textit{failure}, or \textit{refinement}. If \textit{success}, the constraint is accepted, and the task proceeds. If \textit{failure}, the next constraint is evaluated. If \textit{refinement}, the system enters the refinement phase for further optimization.
In the refinement phase, the system performs fine-grained resampling around the predicted interaction direction \( \mathbf{v}_i \) to correct misalignments between the functional and geometric axes of objects. The system uniformly samples six refined directions \( \mathbf{v}_i^{(j)} \) around \( \mathbf{v}_i \) and evaluates them.

\noindent\textbf{Closed-loop Execution.}
Once the interaction primitives and the corresponding spatial constraints \( \mathcal{C} \) are defined for each stage, the task execution can be formulated as an optimization problem. The objective is to minimize the loss function to determine the target pose \( \mathbf{P}^{ee*} \) of the end-effector. The optimization problem can be expressed as:

\vspace{-10pt}
\begin{equation}
\begin{split}
\mathbf{P}^{ee*} = \arg \min_{\mathbf{P}^{ee}} \left\{ \sum_{j=1}^{N} \mathcal{L}_j(\mathbf{P}^{ee}) \right\}, \quad \text{s.t.} \\
\mathcal{L} = \{ \mathcal{L}_C, \mathcal{L}_{\text{collision}}, \mathcal{L}_{\text{path}} \},
\end{split}
\end{equation}

where the constraint loss \( \mathcal{L}_C \) ensures that the action adheres to the task's spatial constraints \( \mathcal{C} \), and is defined as

\vspace{-10pt}
\begin{equation}
\mathcal{L}_C = \rho(\mathcal{C}, \mathbf{P}_t^{\text{active}}, \mathbf{P}_t^{\text{passive}}), \; \text{where} \;
\mathbf{P}_t^{\text{active}} = \Phi(\mathbf{P}_t^{ee})
\label{equation:ee_slover}
\end{equation}

Here, \( \rho(\cdot) \) measures the deviation between the current spatial relationship of the active object \( \mathbf{P}_t^{\text{active}} \) and the passive object \( \mathbf{P}_t^{\text{passive}} \) from the desired constraint \( \mathcal{C} \), while \( \Phi(\cdot) \) maps the end-effector pose to the active object's pose. The collision loss \( \mathcal{L}_{\text{collision}} \) prevents the end-effector from colliding with obstacles in the environment and is defined as

\vspace{-15pt}
\begin{equation}  
\mathcal{L}_{\text{collision}} = \sum_{j=1}^{N} \max \left( 0, d_{\text{min}} - d(\mathbf{P}^{ee}, \mathbf{O}_j) \right)^2,
\label{equation:collision_loss}
\end{equation}
\vspace{-10pt}

where \( d(\mathbf{P}^{ee}, \mathbf{O}_j) \) represents the distance between the end-effector and the obstacle \( \mathbf{O}_j \), and \( d_{\text{min}} \) is the minimum allowable safety distance. The path loss \( \mathcal{L}_{\text{path}} \) ensures smooth motion and is defined as

\vspace{-10pt}
\begin{equation}
\mathcal{L}_{\text{path}} = \lambda_1 d_{\text{trans}}(\mathbf{P}_t^{ee}, \mathbf{P}^{ee}) + \lambda_2 d_{\text{rot}}(\mathbf{P}_t^{ee}, \mathbf{P}^{ee}),
\end{equation}

where \( d_{\text{trans}}(\cdot) \) and \( d_{\text{rot}}(\cdot) \) represent the translational and rotational displacements of the end-effector, respectively, and \( \lambda_1 \) and \( \lambda_2 \) are weighting factors that balance the influence of translation and rotation. By minimizing these loss functions, the system dynamically adjusts the end-effector pose \( \mathbf{P}^{ee} \), ensuring successful task execution while avoiding collisions and maintaining smooth motion.

While~\Eref{equation:ee_slover} outlines how interaction primitives and their corresponding spatial constraints can be leveraged to optimize the executable end-effector pose, real-world task execution often involves significant dynamic factors. For instance, deviations in the grasp pose may result in unintended object movement during a grasping task. Moreover, in certain dynamic environments, the target object may be displaced. These challenges highlight the critical importance of closed-loop execution in handling such uncertainties.
To address these challenges, our system leverages the proposed object-centric interaction primitives and directly employs an off-the-shelf 6D object pose tracking algorithm to continuously update the poses of both the active object \( \mathbf{P}_t^{\text{active}} \) and the passive object \( \mathbf{P}_t^{\text{passive}} \) in real-time, as required in~\Eref{equation:collision_loss}. This real-time feedback allows for dynamic adjustments to the target pose of the end-effector, enabling robust and accurate closed-loop execution.

\vspace{-5pt}
\section{Experiment}

In this section, we aim to answer the following questions: \textbf{(1)} To what extent does OmniManip perform effectively in open-vocabulary manipulation tasks across diverse real-world scenarios (\Sref{sec:quantitative_results})? \textbf{(2)} What role do the system’s critical features play in enhancing its overall performance (\Sref{sec:key_features})? \textbf{(3)} How promising is OmniManip for automating the collection of robot manipulation trajectories to enable scalable imitation learning (\Sref{sec:data_gen_for_IL})?

\subsection{Experimental Setup}
\noindent\textbf{Hardware Configuration.}
Our experimental platform is built around a Franka Emika Panda robotic arm, with its parallel gripper's fingers replaced by UMI fingers\cite{umi}. For perception, we employ two Intel RealSense D415 depth cameras. One camera is mounted at the gripper to provide a first-person view of the manipulation area, while the second camera is positioned opposite the robot to offer a third-person view of the workspace. 

\noindent\textbf{Tasks and Metrics.}
As shown in~\Fref{fig:teaser}, We designed 12 tasks to evaluate models' manipulation capabilities in real-world scenarios. Six of these involve rigid object manipulation (\eg, \emph{pour tea}), while the others focus on articulated manipulation (\eg, \emph{open the drawer}). These tasks cover a diverse set of objects and are intended to assess the models' ability to generalize and adapt in complex environments.
For each task, 10 trials were performed for each approach, and the success rate was recorded. After each trial, the object layout was reconfigured to ensure robust evaluation.

\noindent\textbf{Baselines.}
We compare our approach with three baselines: 1) \textbf{VoxPoser}\cite{voxposer}, which uses LLM and VLM to generate 3D value maps for synthesizing robot trajectories, excelling in zero-shot learning and closed-loop control; 2) \textbf{CoPa}\cite{copa}, which introduces spatial constraints of object parts and combines with VLM to enable open-vocabulary manipulation; and 3) \textbf{ReKep}\cite{rekep}, which employs relational keypoint constraints and hierarchical optimization for real-time action generation from natural language instructions.

\noindent\textbf{Implement Details}
We use GPT-4O from OpenAI API as the vision-language model, leveraging a small set of interaction examples as prompts to guide the model's reasoning for manipulation tasks. The specific prompts used are detailed in the appendix. We employ off-the-shelf models \cite{gsnet, anygrasp} for 6-DOF universal grasping and utilize GenPose++\cite{omni6dpose} for universal 6D pose estimation.

\subsection{Open-Vocabulary Manipulation}
\label{sec:quantitative_results}

We conducted a comprehensive evaluation of OmniManip on 12 open-vocabulary manipulation tasks, ranging from straightforward actions such as pick-and-place to more complex tasks involving object-object interactions with directional constraints and articulated object manipulation. As shown in ~\Tref{table:main_table}, our method exhibits robust zero-shot generalization and superior performance across the board without task-specific training. This generalization capability can be attributed to the commonsense knowledge embedded in VLM, while the proposed efficient object-centric interaction primitives facilitate precise 3D perception and execution. Additionally, we provide qualitative results in the appendix.
OmniManip exhibits a substantial performance advantage over baseline methods, primarily due to two key factors: 1) the efficiency and stability of the proposed object-centric canonical interaction primitives, as further validated through extensive experiments in~\Sref{sec:key_features},  and 2) the advanced dual closed-loop system for planning and execution.  By incorporating a novel self-correction mechanism based on RRC, the system effectively mitigates hallucination issues of large models.  As shown in~\Tref{table:main_table}, this closed-loop planning yields over a 15\% improvement in performance for both rigid and articulated object manipulation tasks.  A detailed qualitative analysis of the closed-loop reasoning and execution is provided in~\Sref{sec:key_features}.

\begin{table}[t]
    \centering
    \begin{tabular}{cccccc}
\toprule
\textbf{Method}           & $0^\circ$ & $25^\circ$ & $45^\circ$ & $75^\circ$ & $90^\circ$ \\ \midrule \midrule
\textbf{ReKep}            & 0/10    & 1/10     & 3/10     & 5/10     & \textbf{7/10 }    \\
\textbf{OmniManip} & \textbf{7/10}    & \textbf{8/10}     & \textbf{8/10}     & \textbf{7/10}     & \textbf{7/10}     \\ \bottomrule
\end{tabular}
    \vspace{-5pt}
    \caption{Quantitative analysis of the impact of viewpoints on the performance, using \emph{`Recycle the battery'} as a case study.}
    \label{table:viewpoint_consistency_analysis}
    \vspace{-20pt}
\end{table}

\subsection{Core Attributes of OmniManip}
\label{sec:key_features}
\begin{figure}[t]
    \centering
    \includegraphics[width=\linewidth]{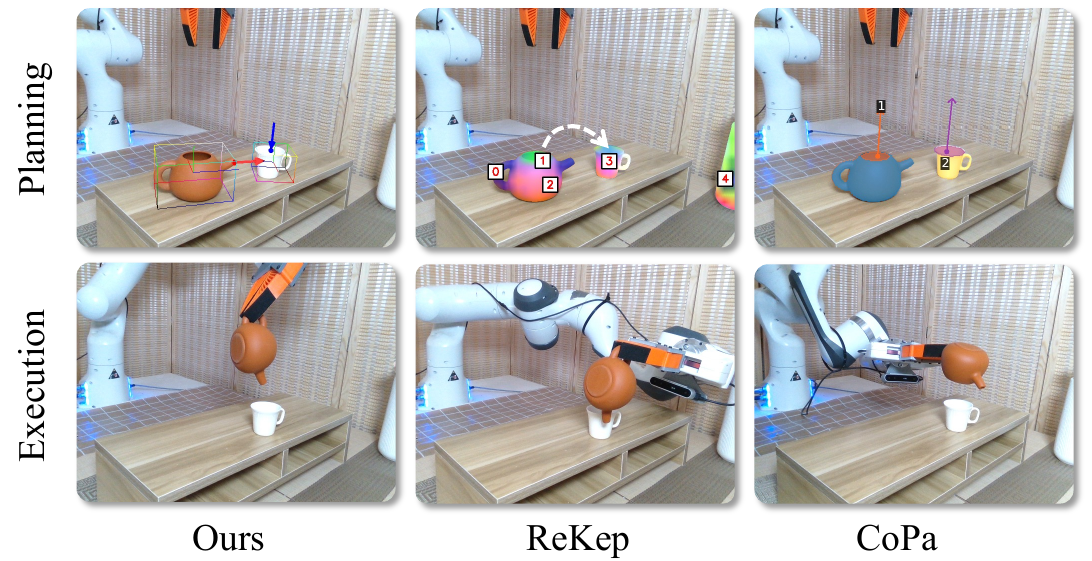} 
    \vspace{-15pt}
    \caption{
    \textbf{Stability analysis of interaction primitives.} 
    Visualization of planning and corresponding execution results across different methods, demonstrated using the `\textit{Pour tea}' as a case study.}
    \label{fig:stability_analysis}
    \vspace{-10pt}
\end{figure}

\begin{figure}[t]
    \centering
    \includegraphics[width=\linewidth]{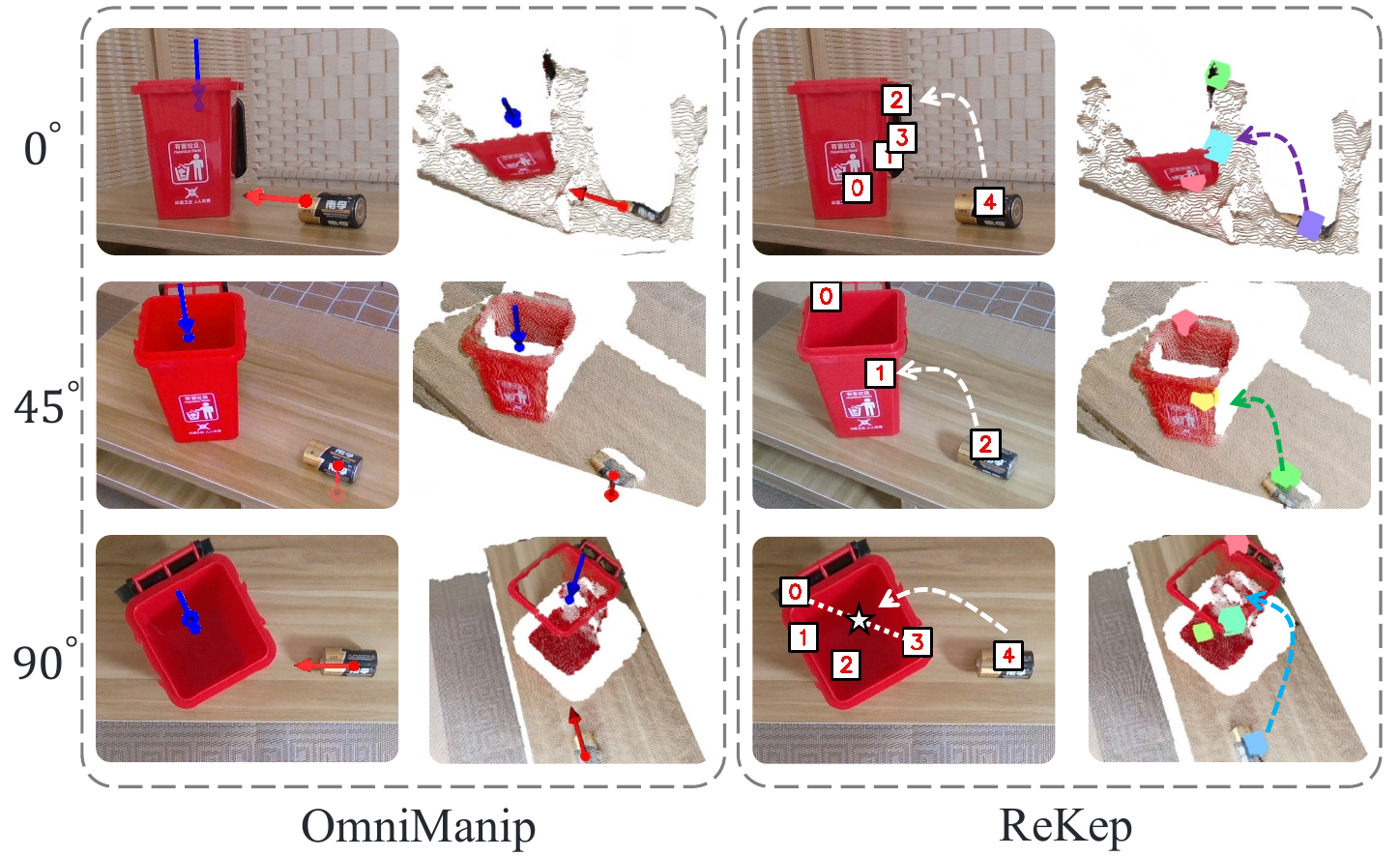} 
    \caption{Qualitative analysis of the impact of viewpoints on the performance, using \emph{`Recycle the battery'} as a case study.}
    \label{figure:viewpoint_consistency_analysis}
    \vspace{-15pt}
\end{figure}

\noindent\textbf{Reliability of OmniManip.}
To effectively bridge VLM with low-level manipulation, reliable interaction primitives are crucial. We evaluate this across two key dimensions: stability and viewpoint consistency.
\textbf{Stability} indicates the reliable extraction of task-relevant interaction primitives. As shown in~\Fref{fig:stability_analysis}, ReKep extracts keypoint proposals through semantic clustering but lacks sensitivity to spatial geometry and task, making it challenging to generate sufficient task-relevant keypoints. CoPa extracts parts via explicit pixel segmentation, exhibiting high sensitivity to image texture and part shape. In contrast, OmniManip, an object-centric interaction primitive, samples interaction points in a canonical space aligned with the object’s functionality, ensuring both robustness and task-specific precision.
\textbf{Consistency} of primitive extraction across varying viewpoints is critical to ensuring the stability of manipulation. Both ReKep and CoPa exhibit difficulties in this regard due to their reliance on sampling points directly from the object's surface. Taking ReKep as an example, ~\Fref{figure:viewpoint_consistency_analysis} illustrates the planning results of ReKep and OmniManip for the `\textit{Recycle battery}' task across different viewpoints. As shown, ReKep successfully identifies interaction points from a 90$^\circ$ top-down view but fails under a 0$^\circ$ frontal view, where the ideal target point is floating in the air. In contrast, OmniManip utilizes an object-centric primitive representation in a canonical space, ensuring viewpoint invariance. ~\Tref{table:viewpoint_consistency_analysis} presents the quantitative comparison, demonstrating that OmniManip's performance is nearly invariant across varying viewpoints, whereas ReKep's performance is significantly affected by changes in viewpoint.

\begin{table}[h]
    \centering
    \vspace{-5pt}

\begin{tabular}{ccccc}
\toprule
\textbf{Sampling}     & \multicolumn{2}{c}{Recycle Battery} & \multicolumn{2}{c}{Pour Tea} \\ 
\cline{2-5}

 \textbf{Method} & Suc. Rate  & Iter.   & Suc. Rate  & Iter.    \\  \midrule \midrule
\textbf{Uniform}    & 50\% & 1.8    &   30\% & 3.4     \\
\textbf{OmniManip} & \textbf{80\%} & \textbf{1.7}    & \textbf{70\%} & \textbf{1.8 }       \\ \bottomrule
\end{tabular}



    \vspace{-5pt}
    \caption{Quantitative analysis of the primitive sampling efficiency.}
    \label{table:sampling_efficiency}
    \vspace{-8pt}
\end{table}

\noindent\textbf{Efficiency of OmniManip.}
Interaction direction proposals in OmniManip are driven by a targeted sampling strategy. Compared with uniform sampling in SO(3), OmniManip samples along the principal axes of the object’s canonical space. Since the canonical space is aligned with the object's functionality, this ensures both efficient and effective sampling. To evaluate this efficiency, we compared OmniManip's sampling strategy with uniform sampling in SO(3) using two key metrics: the number of iterations and the corresponding task success rate. As shown in~\Tref{table:sampling_efficiency} OmniManip not only requires fewer iterations but also achieves superior task performance, demonstrating that aligning the sampling process with the object's functionality reduces sampling overhead while improving overall performance.

\begin{figure}[t]
    \centering
    \includegraphics[width=\linewidth]{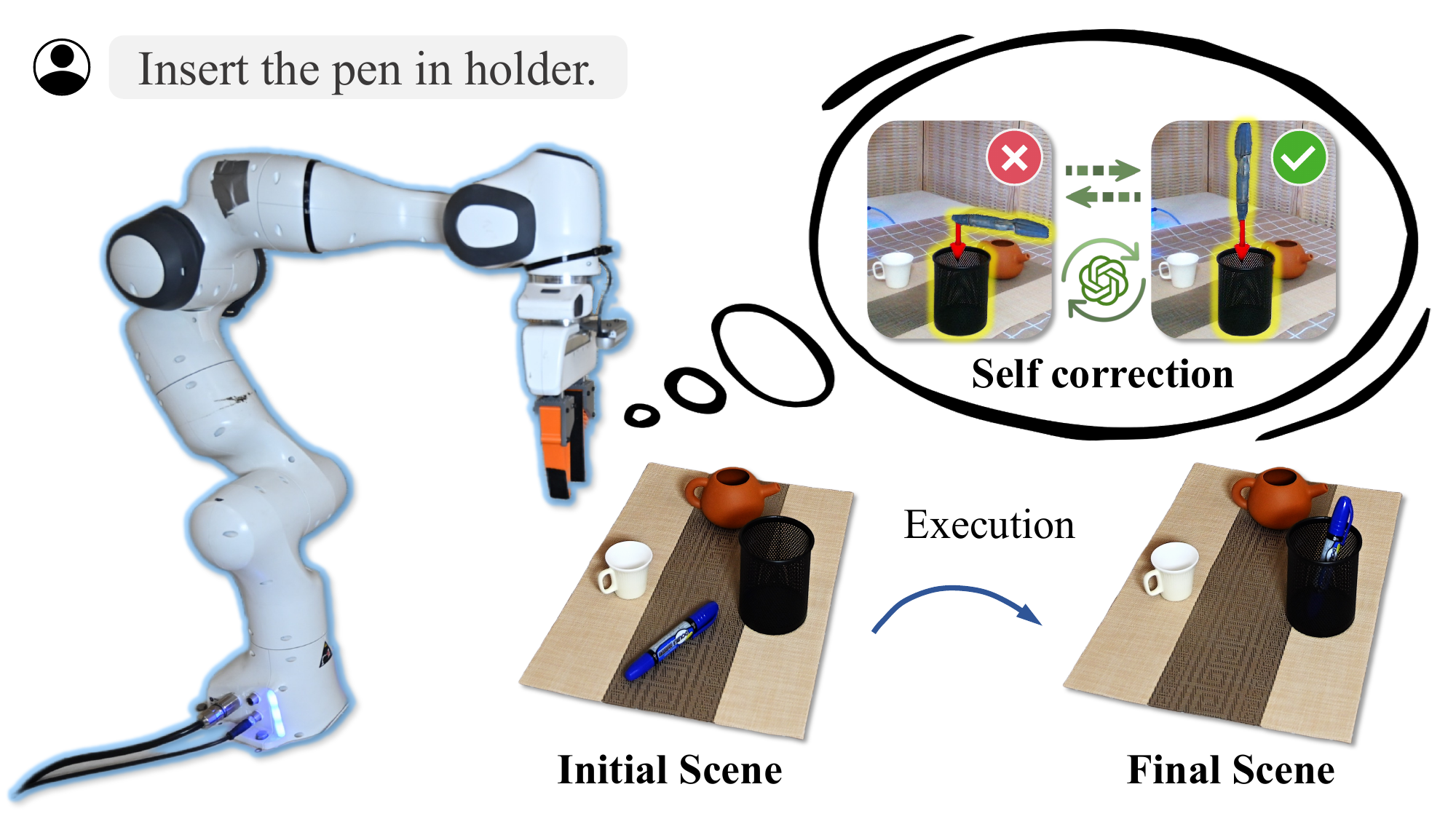} 
    \caption{Closed-planning. Self-correction mechanism via RRC.}
    \label{fig:closed-loop_planning}
    \vspace{-15pt}
\end{figure}

\noindent\textbf{Closed-Loop Planning.}
In current methods, the planning component of VLM operates in an open-loop manner, meaning it cannot verify the correctness of the plan before execution. While ReKep achieves closed-loop control through point tracking, this only functions at the execution stage and does not provide feedback on the planning results generated by the VLM. In contrast, OmniManip introduces a unique self-correction mechanism via RRC, achieving closed-loop planning, which significantly reduces planning failures caused by VLM hallucinations, thereby offering more reliable planning. We report the results with closed-loop planning disabled in~\Tref{table:main_table}, where the task success rate decreases by over 15\% in both rigid and articulated object manipulation tasks, demonstrating the effectiveness of the closed-loop planning approach. In~\Fref{fig:closed-loop_planning}, we qualitatively illustrate the closed-loop planning results using the "Insert the pen in a holder" task as an example. It is evident that OmniManip can effectively pre-render the planning outcomes and achieve self-correction through the RRC process, thereby enabling closed-loop planning.

\begin{table}[h]
    \centering
    \begin{tabular}{lc}
\toprule
\textbf{Task}     &  Success Rate \\ \midrule \midrule
\textbf{Pick up the cup on the dish}     & 95.24\%      \\
\textbf{Recycle the battery} & 91.30\%         \\ 
\textbf{Insert the pen in holder}  & 86.36\%  \\

\bottomrule
\end{tabular}

    \caption{
    Behavior cloning with demonstrations from OmniManip.
    }
    \label{table:il_policy}
    \vspace{-4pt}
\end{table}

\begin{figure}[t]
    \centering
    \includegraphics[width=\linewidth]{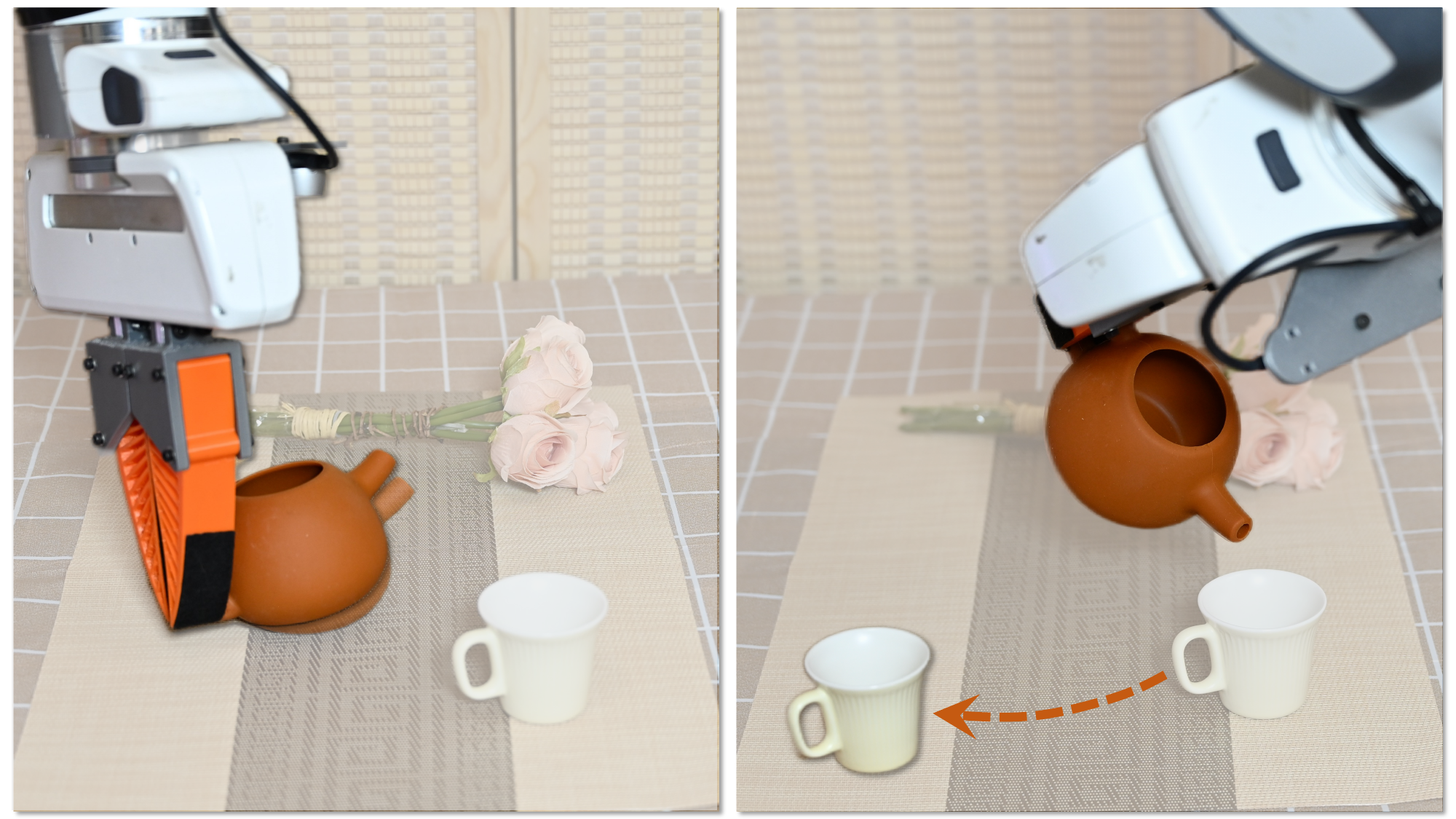} 
    \caption{Two typical failure cases without closed-loop execution. 
    }
    \label{fig:closed-loop_execution}
    \vspace{-15pt}
\end{figure}

\noindent\textbf{Closed-Loop Execution.}
Even with perfect planning, open-loop execution can still lead to task failure. ~\Fref{fig:closed-loop_execution} illustrates two typical examples where planning succeeds, but open-loop execution causes failure. In the left image of~\Fref{fig:closed-loop_execution}, the relative pose between the gripper and the object changes during the interaction, while the right image of~\Fref{fig:closed-loop_execution} shows a scenario where the target pose is dynamic, such as when the object moves during the task. To address these challenges, OmniManip employs pose tracking to enable real-time closed-loop execution. Recent work, ReKep, uses point tracking for closed-loop control but suffers from occlusions, leading to a 47\% failure rate \cite{rekep}. In contrast, OmniManip demonstrates greater robustness to occlusions caused by object movement. This is a benefit of object-centric pose tracking, enabling continued tracking of canonical space interaction primitives based on the object pose, even when the primitives are no longer visible.

\subsection{OmniManip for Demonstration Generation}
\label{sec:data_gen_for_IL}

We employed OmniManip to generate automatic demonstration data. Unlike prior methods reliant on task-specific privileged information, OmniManip collects demonstration trajectories for new tasks in a zero-shot manner, without needing task-specific details or prior object knowledge. To validate the effectiveness of OmniManip-generated data, we collected 150 trajectories per task to train behavior cloning policies \cite{diffusion_policy}. These policies achieved high success rates, as shown in Table \ref{table:il_policy}. Additional tasks and detailed results are provided in the appendix.
\section{Conclusion}
In this work, we presented a novel object-centric intermediate representation that effectively bridges the gap between VLM and the precise spatial reasoning required for robotic manipulation. We structured interaction primitives in object canonical space to translate high-level semantic reasoning into actionable 3D spatial constraints. The proposed dual closed-loop system ensures robust decision-making and execution, all without VLM fine-tuning. Our approach demonstrates strong zero-shot generalization across a variety of manipulation tasks, highlighting its potential for automating robotic data generation and improving the efficiency of robotic systems in unstructured environments. This work provides a promising foundation for future research into scalable, open-vocabulary robotic manipulation.
\noindent{\textbf{Limitations.}}
While advantageous, OmniManip also has limitations. It cannot model deformable objects due to pose representation. Its effectiveness also hinges on the mesh quality of 3D AIGC, which remains challenging despite progress. Additionally, multiple VLM calls present computational challenges, even with parallel processing.
\section*{Acknowledgments}
We would like to thank Mingdong Wu and Tianhao Wu from PKU for their fruitful discussions, and Baifeng Xie from AgiBot for valuable technical support.

{
    \small
    \bibliographystyle{ieeenat_fullname}
    \bibliography{main}
}
\end{document}